\documentclass[pdflatex, sn-mathphys-num]{sn-jnl}

\usepackage{graphicx}%
\usepackage{multirow}%
\usepackage{amsmath,amssymb,amsfonts}%
\usepackage{amsthm}%
\usepackage{mathrsfs}%
\usepackage[title]{appendix}%
\usepackage{xcolor}%
\usepackage{textcomp}%
\usepackage{manyfoot}%
\usepackage{booktabs}%
\usepackage{algorithm}%
\usepackage{algorithmicx}%
\usepackage{algpseudocode}%
\usepackage{listings}%

\usepackage{hyperref}
\usepackage{bookmark} 
\usepackage{anyfontsize} 

\usepackage{tabularx}
\usepackage{booktabs}
\usepackage{makecell}
\usepackage{adjustbox}

\usepackage{doi}

\theoremstyle{thmstyleone}%
\theoremstyle{thmstyletwo}%

\theoremstyle{thmstylethree}%

\begin{document}

\title[Article Title]{Introducing Environmental Constraints to Grasping Strategies for Paper-Like Flexible Materials Using a Soft Gripper}


\author[1,2]{\fnm{Yi} \sur{Dong}} 

\author[1,2]{\fnm{Yang} \sur{Li}} 

\author*[1,2]{\fnm{Jinjun} \sur{Duan}}\email{duan-jinjun@nuaa.edu.cn}

\author*[1,2]{\fnm{Zhendong} \sur{Dai}}\email{zddai@nuaa.edu.cn}

\affil*[1]{\orgdiv{College of Mechanical and Electrical Engineering}, \orgname{Nanjing University of Aeronautics and Astronautics}, \orgaddress{\street{29 Yudao Street}, \city{Nanjing}, \postcode{211106}, \state{Jiangsu}, \country{China}}}

\affil[2]{\orgdiv{Jiangsu Key Laboratory of Bionic Materials and Equipment}, \orgname{Nanjing University of Aeronautics and Astronautics}, \orgaddress{\street{29 Yudao Street}, \city{Nanjing}, \postcode{211106}, \state{Jiangsu}, \country{China}}}



\abstract{
Robotic manipulation of flexible objects is widely required in both industrial and service applications. 
Among such objects, paper-like materials exhibit distinct mechanical characteristics compared to cloth, being more sensitive to compressive stress, where minor variations in physical properties can significantly affect grasping.
This study systematically investigates grasping strategies for paper-like materials using a universal soft gripper by exploiting environmental constraints.
Based on manipulation primitives employed in existing grasping strategies, we proposed systematic grasping strategies for flexible materials by exploiting environmental constraints and analyzed their mechanical and kinematic models.
To investigate the influence of materials and working conditions on grasping, an evaluation system for measuring grasping force and success rate was defined and experimentally evaluated. 
Finally, we summarized the specific workspaces and characteristics of different strategies that can satisfy various task requirements and lead to potential applications in household service robots for grasping planar flexible objects.
}




\maketitle

\section{Introduction}\label{sec1}

The ability to reliably grasp and manipulate deformable objects constitutes a core challenge and an enabling skill for robots transitioning from structured industrial settings to the unstructured environments of human society \cite{xu2025model, Wang2024A}.
Although existing studies on cloth  as a flexible material have focused on activities such as folding clothes and making beds, paper and cloth have different physical properties. 
Because of its soft and rigid characteristics, paper is one of the most complex objects for robotic manipulation \cite{namiki2015robotic, namiki2021origami}.
The viscoelasticity of different types of papers varies significantly. 
In addition, folding can change the properties of paper because of the formation of crease lines or wrinkles. Consequently, it can easily be torn when grasped with a rigid gripper.

To address the infinite degrees of freedom and stiffness of paper, different grasping mechanisms have been proposed. 
The suction cups developed in \cite{wang2023research, koivikko20213d, chin2020multiplexed} were used to manipulate thin and soft objects owing to  their nondestructive, evenly distributed grasping force and simple grasping procedure. 
However, suction cups are mainly used in industrial logistics, where tasks are limited to simple pick-and-place operations, and are therefore less suitable for home service scenarios that involve complex secondary manipulation tasks and a wide variety of objects.
Complex robotic systems, such as specially designed robot manipulators \cite{balkcom2008robotic}, multi-arm robots \cite{Tanaka2007Origami}, and multi-fingered hands \cite{elbrechter2012folding, namiki2015robotic} have been employed for grasping paper-like materials.
These systems are capable of grasping and folding papers and even completing complex origami tasks. However, they are not suitable for practical applications because of their complexity and high cost.
In recent years, paper-grasping strategies using soft grippers have been proposed \cite{jiang2019dynamic}, and these grippers have been proven to exhibit superior multiple compliance when handling thin and flexible objects \cite{teeple2022multi}.
A soft gripper can also interact compliantly with environmental constraints, which can constrain a paper with multiple degrees of freedom.
Therefore, this study  focused on a grasping strategy using soft grippers and environmental constraints for grasping paper-like materials.

Existing studies on paper have focused on grasping \cite{jiang2024rotipbot, xiong2023rapid, yoshimi2012picking, hirai1994modeling, wakamatsu2004static}, origami \cite{namiki2021origami, namiki2015robotic, elbrechter2012folding, balkcom2008robotic, Tanaka2007Origami}, orienting \cite{kristek2012orienting, srinivasa2002experiments, mason1999mobile}, flipping \cite{zhao2023flipbot, zheng2022autonomous, jiang2019dynamic}, and shape control \cite{tong2023deep, guo2021deformation}. Among these tasks, grasping is the most basic step and serves as a necessary step in subsequent operations.
Previous studies on grasping have focused on developing an analytical method for modeling the shape of thin bendable objects \cite{wakamatsu2004static, hirai1994modeling}. This method aids in planning manipulation operations for deformable parts.
However, these studies discuss only a top-grasping strategy, with no further in-depth discussion on grasping position and paper specifications.
Existing studies on paper-grasping strategies have focused on specially designed grippers, such as grippers with nails \cite{yoshimi2012picking}, those with biomimetic bistable elements \cite{xiong2023rapid}, and those with rotatable tactile fingertips \cite{jiang2024rotipbot}.
These strategies improve the success rate and efficiency of paper grasping; however, they are direct-grasping strategies developed for specific grippers.
In our previous work \cite{dong2023robotic}, we proposed several grasping strategies based on a soft gripper; however, that study was not comprehensive. 
In particular, it did not include the top-scooping strategy considered in this paper, and the strategies were introduced only qualitatively, without a systematic investigation of their underlying mechanisms, applicable conditions, workspaces, or working characteristics.
This research explores the systematic grasping strategies of a universal soft gripper within a gripper--object--environment system, testing its grasping workspaces and characteristics.

In this work, we systematically investigate grasping strategies for paper-like materials using a universal soft gripper and environmental constraints, analyzing their working characteristics and providing a reference for robotic grasping of flexible materials.
The main contributions of this study are as follows:

1) Systematic grasping strategies that utilize environmental constraints for grasping paper-like materials are proposed. 
By investigating existing direct top-grasping methods, additional environmental constraints are introduced into the grasping of flexible materials, such as walls and edges. 
The grasping processes of these strategies are elaborated in detail, and their mechanical and kinematic models are analyzed.
This contribution extends and deepens the understanding of grasping strategies for thin flexible materials.

2) The workspaces and properties of these strategies are evaluated and summarized.
The grasping force and success rate in the actual grasping process are tested, the working space of these strategies for different specifications of paper is explored, and their working characteristics are summarized for reference in related studies.
These results provide directly applicable guidance for grasping paper-like and other everyday thin flexible materials.

The remainder of this paper is organized as follows:
Related studies are discussed in Section \ref{RELATED WORK}. 
In Section \ref{GRASPING STRATEGIES}, we summarize all grasping strategies that exploit environmental constraints and their grasping models.
In Section \ref{SETUP FOR EXPERIMENTS}, we describe the experimental settings, including the robotic system architecture, strategy algorithm, evaluation system, and specimen.
In Section \ref{EXPERIMENTAL RESULTS}, we present and analyze the experimental results of different strategies under different working conditions.
The discussion and conclusions are presented in Sections \ref{DISCUSSION} and \ref{CONCLUSION}, respectively.

\section{Related Work}
\label{RELATED WORK}

\subsection{ Grasping of Thin Flat Objects }
Currently, the grasping methods for thin objects can be classified into three groups: those using specialized strategies \cite{zhang2022prying, he2021scooping, babin2019stable, babin2018picking, odhner2013open, odhner2012precision}, customized grippers \cite{yuan2020design, morino2020sheet, ko2020tendon}, or environmental constraints \cite{ding2024preafford, hang2019pre, bimbo2019exploiting, sarantopoulos2018human}.
Specialized strategies include prying \cite{zhang2022prying} and scooping \cite{he2021scooping, babin2019stable, odhner2013open, odhner2012precision}, with the difference that prying is mainly achieved by standard grippers using motion control, whereas scooping requires fingers with sharper fingertips to facilitate insertion under the object.
In addition, scooping is usually achieved in combination with compliant or underactuated grippers \cite{babin2019stable, babin2018picking, odhner2013open, odhner2012precision}, thus simplifying the control of contact forces.
The most significant feature of customized grippers is that they have movable contact surfaces \cite{he2021scooping, yuan2020design, morino2020sheet, ko2020tendon}, which are convenient for picking up thin objects with small contact areas. 
The last method involves the use of environmental constraints for completing pre-grasping, thus facilitating the final grasping \cite{ding2024preafford, hang2019pre, bimbo2019exploiting, sarantopoulos2018human}. This method will be discussed in detail in Section \ref{Exploiting environmental constraints}. 
The aforementioned studies have extensively investigated grasping strategies for thin objects. However, most of them focused on rigid objects.
Current methods for grasping flexible objects rely on customized grippers and are mainly used for verifying the functionality of these grippers rather than investigating their grasping strategies in depth.
This study focused on the grasping of flexible materials and investigated grasping strategies for flexible materials using general soft grippers.

\subsection{Paper Grasping and Manipulation}

Studies on the grasping of flexible objects began around the 21st century. 
These studies aimed at developing an analytical method for modeling the shape of a deformable object for easy manipulation \cite{hirai1994modeling, wakamatsu2004static}. 
However, they focused only on the top-grasping strategy and did not explore the grasping model or working characteristics of the strategy.
Mason et al. \cite{mason1999mobile, srinivasa2002experiments} developed a mobipulator, comprising a small car with independently powered wheels, to manipulate paper on a desk surface. 
This method is effective as it combines locomotion and manipulation; however, the use of cars as mobile platforms limits their applicability to tasks requiring precise manipulation or handling of a variety of objects.
From 2010, studies on origami robots that  can fold different types of interesting artworks, such as paper airplanes and cranes, began being conducted \cite{namiki2015robotic, elbrechter2012folding, balkcom2008robotic, Tanaka2007Origami}.
However, these tasks focused on task planning rather than grasping strategies or the use of complex robotic systems.
Several studies on paper manipulation have been conducted, focusing on aspects such as grasping \cite{yoshimi2012picking, xiong2023rapid, jiang2024rotipbot}, reorientation \cite{kristek2012orienting}, flipping \cite{jiang2019dynamic, zheng2022autonomous, zhao2023flipbot}, and shape control \cite{guo2021deformation, tong2023deep}.
Among these aspects, we focused on paper-grasping strategies. These methods are implemented from a hardware perspective, using specially designed grippers, including grippers with nails \cite{yoshimi2012picking}, those with bistable elements \cite{xiong2023rapid}, and those with rotatable fingertips \cite{jiang2024rotipbot}.
Using a universal soft gripper and existing grasping methods, we proposed systematic grasping strategies for paper-like materials and conducted  in-depth research on these strategies.

\subsection{Exploiting Environmental Constraints} 
\label{Exploiting environmental constraints}

Environmental constraints in robotics, which have long been exploited by humans during daily manipulation, include tabletops, walls, and edges \cite{dafle2014extrinsic, dong2023robotic, sundaram2020environment, eppner2015planning, eppner2015exploitation}.
These constraints can assist robots in grasping \cite{turco2021grasp, bimbo2019exploiting, sarantopoulos2018human}, orienting \cite{chavan2018regrasping, almeida2017dexterous, chavan2020sampling, chavan2015prehensile, wan2015reorientating}, pivoting \cite{tong2020picking, fakhari2021motion, hou2018fast, holladay2015general}, or rolling \cite{specian2018robotic} objects.
These constraints have unique advantages in grasping difficult-to-grasp objects, such as flat objects \cite{ding2024preafford, bimbo2019exploiting, hang2019pre, sarantopoulos2018human}, hats, and inverted bowls.
The use of environmental constraints is an efficient and convenient method for grasping planar rigid objects; studies using these methods currently focus on pre-grasping planning \cite{bimbo2019exploiting, sarantopoulos2018human} and learning \cite{ding2024preafford, hang2019pre}.
Moreover, environmental constraints have been implicitly used for grasping planar flexible objects \cite{borras2020grasping}.
For example, contact with a tabletop limits the shapes of flexible objects, similar to the function of a gripper.
In addition to using tabletops, Shawn et al. \cite{kristek2012orienting} employed dual-arm robots without sensors to orient paper using walls .
However, thus far, no relevant studies have systematically investigated grasping strategies that utilize environmental constraints for grasping flexible materials.
In this study, we introduced environmental constraints into the grasping strategies for flexible materials and thoroughly investigated these strategies.

\section{Robotic Grasping Strategies}
\label{GRASPING STRATEGIES}

\subsection{Primitive Skills for Paper Grasping} 
\label{Classification of skills}

\begin{figure}[htbp]
  \centering
  \includegraphics[width=0.96\textwidth]{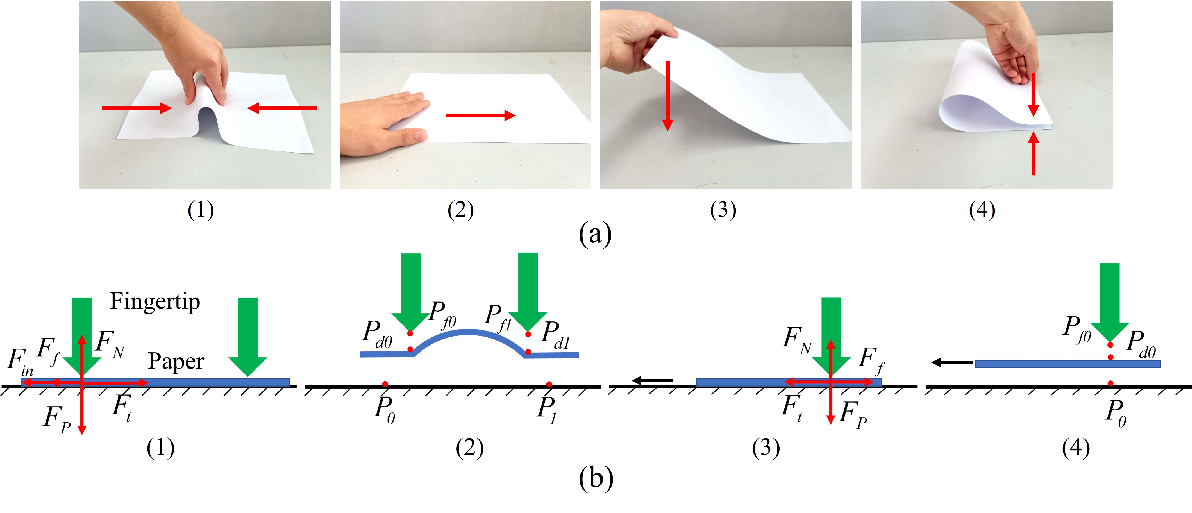} 
  \caption{Human one-handed manipulation primitives and the corresponding gripper-based primitives inspired by them. (a) Human one-handed manipulation primitives for paper, including pinching, sliding, placing, and folding. (b) Paper manipulation skills: (1) Force analysis of the pinch skill, (2) Description of pinching motion, (3) Force analysis of the slide skill, (4) Description of sliding motion}
  \label{skillforce}
\end{figure}

Similar to the primitives for manipulating fabrics with one hand \cite{shibata2009wiping}, those for manipulating paper include pinching, sliding, placing, and folding (Fig. \ref{skillforce} (a)).
The pinch and slide operations are key primitives for human grasping of objects.
Based on these primitives, pinching and sliding skills were developed for soft grippers (Fig. \ref{skillforce} (b)).

We established mechanical and kinematic models for the pinch primitive (Figs. \ref{skillforce}(b) (1) and (2)), which involves deformation and sliding of the paper.
For mechanical analysis (Fig. \ref{skillforce}(b) (1)), let $F_P$ be the pressing force exerted by the gripper’s finger, $F_N$ is the normal force, which is equal to $F_P$, $F_\tau$ is the tangential force exerted by the gripper’s finger, $F_f$ is the friction force between the paper and table, and $F_{in}$ is the resistance to deformation from the paper. 
$\mu_0$ is the coefficient of friction between the fingertip and  paper, and $\mu_1$ is the coefficient of friction between the paper and table.
For kinematic analysis (Fig. \ref{skillforce}(b) (2)), assuming the contact between a finger and paper is considered a point contact. 
When a finger pinches the paper, $P_{fi}$ denotes the contact point on the $i$th fingertip, $P_{di}$ denotes the contact point on the surface of the deformable paper, and $P_i$ denotes the point fixed to  the floor during displacement.

The pinching skill comprises two stages: 
The first stage is the contact stage, in which  no slip occurs.
At this moment, $F_\tau$ is less than the sum of $F_{in}$ and the maximum friction force between the paper and table. 
Therefore, the object remains at rest, and the first stage can be described by Eqs. (\ref{eq4}) and (\ref{eq5}): 
\begin{equation}\label{eq4}
    F_f + F_{in} = F_{\tau}
\end{equation}
\begin{equation}\label{eq5}
    \forall i ( P_{fi} = P_{di} = P_i)
\end{equation}

In Eq. (\ref{eq4}), $F_f \leq F_N \mu_1$, and $F_{in}$  is close to zero because no noticeable deformation is observed at this time. 
Therefore, the necessary condition for Eq. (\ref{eq4}) is $F_{\tau}\leq F_N \mu_1$. 
This implies that the active resultant force exerted by the gripper is within the friction cone of the table and paper.

The second stage is the movement stage, during which the paper begins to deform.
At this time, $F_\tau$ exceeds the sum of $F_{in}$ and the maximum friction force between the paper and table. 
The second stage can be described by Eqs. (\ref{eq6}) and (\ref{eq7}).
\begin{equation}\label{eq6}
    F_f + F_{in} < F_{\tau}
\end{equation}
\begin{equation}\label{eq7}
    \forall i ( P_{fi} = P_{di}) \land{\exists i ( P_{di} \neq P_i)}
\end{equation}

In Eq. (\ref{eq6}), $F_f = F_N \mu_1$, $F_{in} > 0$, and $F_{\tau} < F_N \mu_0$. Therefore, the criterion for Eq. (\ref{eq6}) can be expressed as Eq. (\ref{eq01}), which is also the initial necessary condition for grasping the paper from the table.

\begin{equation}\label{eq01}
    \mu_1 < \mu_0
\end{equation}
\begin{equation}\label{eq02}
    F_{\tau} - F_f > \frac{\pi^2EI}{(\mu L)^2}
\end{equation}
\begin{equation}\label{eq03}
    F_N(\mu_0 - \mu_1) > \frac{\pi^2EI}{(\mu L)^2}
\end{equation}

When the tangential force $F_{\tau}$ further increases, the external force exerted on the paper exceeds Euler’s critical load (Eq. (\ref{eq02})).
At this moment, the paper continues to deform and enters the yielding stage, enabling the gripper to grasp the wrinkles in the paper.
In this equation, $L$ is the length of the deformed part of the paper; $\mu$ is the length factor, which is determined by the constraints of the deformed part; and $EI$ is the bending stiffness of the piece, where $E$ is Young’s modulus and $I$ is the moment of inertia of the section and is proportional to the third power of the thickness $h$.
In Eq. (\ref{eq02}), $F_{\tau} < F_N \mu_0$ and $F_f = F_N \mu_1$. 
Therefore, the normal force criterion is expressed by Eq. (\ref{eq03}).

The Euler buckling model in Eqs. (\ref{eq02}) and (\ref{eq03}) assumes that the paper behaves as a linear elastic slender column under axial compression with idealized boundary conditions. 
This model is applicable to thin paper-like materials, where the deformed length $L$ is much larger than the thickness (e.g., GSM $\leq$ 250 g in our experiments). 
However, a key limitation is that paper exhibits viscoelastic and plastic behaviors that are not captured by linear elasticity; thus, the model provides a conservative first-order estimate instead of an exact prediction.
The length factor $\mu$ in Eqs. (\ref{eq02}) and (\ref{eq03}) reflects the boundary conditions of the buckled paper segment. 
It is determined by the grasping position relative to the paper edge (Fig. \ref{graspstrategy}). 
When grasping near an edge (edge grasping), the boundary condition approximates a fixed-pinned configuration, thus yielding $\mu \approx 0.7$. 
When grasping far from the edge (non-edge grasping), the condition approximates a fixed-fixed configuration, thus yielding $\mu \approx 0.5$. 
In this study, $\mu$ is regarded as a constant for each grasping configuration based on the dominant boundary condition.

The slide skill involves at least one finger making contact with the paper and moving it, as shown in Figs. \ref{skillforce}(b) (3)and (4).
When displacing the paper, no deformation or slippage occurs between the gripper and paper.
Therefore, except $F_{in} = 0$, the force analysis is similar to that of  the pinching skill (Eq. (\ref{eq8}), and $\mu_1 < \mu_0$ is a necessary condition. 
During displacement, no relative movement is observed between the fingers, as expressed by Eq. (\ref{eq9}).

\begin{equation}\label{eq8}
    F_f < F_{\tau}
\end{equation}
\begin{equation}\label{eq9}
    \lvert P_{di} - P_{dj} \rvert = \lvert P_{fi} - P_{fj} \rvert = const
\end{equation}

\subsection{Grasping by Exploiting Tabletop}
\label{Grasping Exploiting Tabletop}

Based on existing studies on paper grasping, we proposed two direct-grasping strategies, including top grasping and top scooping, using a desktop. Both strategies are implemented based on pinching skills.

\begin{figure}[htbp]
  \centering
  \includegraphics[width=0.96\textwidth]{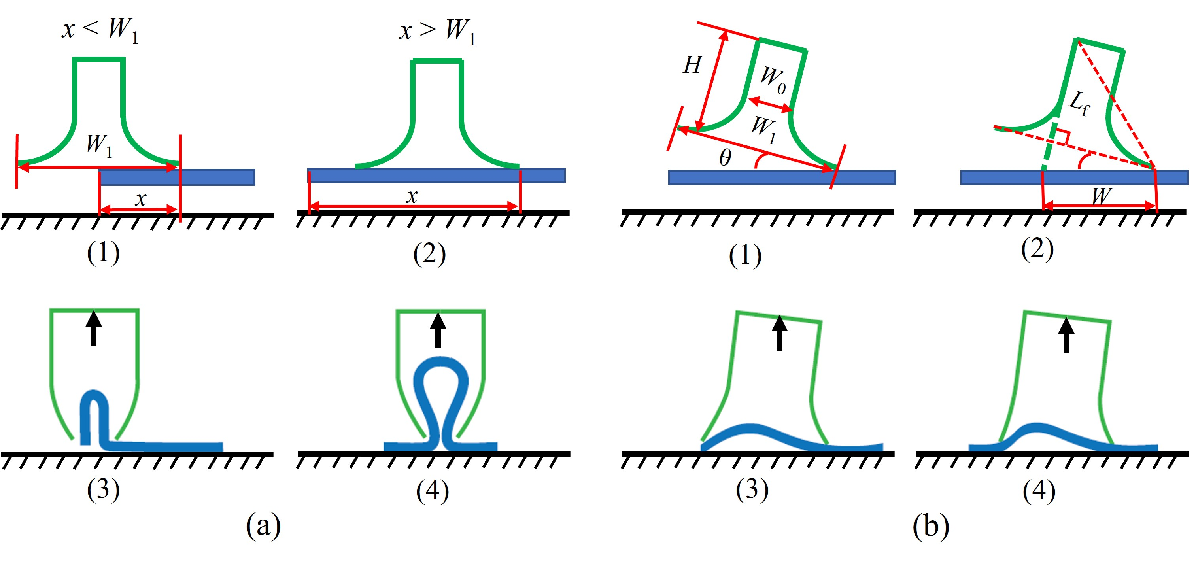}
  \caption{Paper-grasping strategies by exploiting tabletop. (a) Strategy 1: top grasping. (b) Strategy 2: top scooping. The first row shows the different grasping positions of Strategies 1 and 2, and the second row shows the corresponding grasping process of the first row}
  \label{graspstrategy}
\end{figure}

\textbf{Strategy 1} (Figs. \ref{graspstrategy}(a)) is based on the pinch skill and is termed top grasping. 
Although this strategy was proposed as early as \cite{hirai1994modeling}, its focus was on the deformation of thin parts, rather than on the working principles and characteristics of the grasping strategy.
During grasping, the gripper first opens and then approaches and contacts the paper. 
Next, the air pressure in the gripper increases to ensure that the fingers are firmly in contact with the paper and set for grasping.
Finally, when the gripper moves upward, it closes and the paper is grasped automatically.
As discussed in Section \ref{Classification of skills}, the first condition for grasping is satisfied by Eq. (\ref{eq01}), and the tangential and normal forces must satisfy Eq. (\ref{eq02}) and Eq. (\ref{eq03}), respectively.
The initial and maximum gripper openings are denoted by $W_0$ and $W_1$, respectively. 
The grasping position $x$ is classified into two cases: $W_0<x<W_1$ and $W_1<x$ (Fig. \ref{graspstrategy}(a) (1) and (2)), which correspond to edge grasping and non-edge grasping, respectively. 
The length coefficient $\mu$ is associated with the grasping position, and their relationship is described in Section \ref{Classification of skills}.
Figs. \ref{graspstrategy}(a) (3) and (4) show the shapes of the paper after grasping under different constraints in Strategy 1.

\textbf{Strategy 2} involves a variation in the pinching skill, similar to scooping \cite{babin2019stable, babin2018picking} and is therefore termed top scooping.
Its working process is shown in Figs. \ref{graspstrategy}(b). 
Compared with Strategy 1, it introduces a tilt angle at the gripper. 
First, the fingers on one side of the gripper approach and touch the paper. 
After receiving a closing command, the fingers perform a scooping action in the air and make close contact with the paper. 
Finally, as the gripper moves upward, it automatically grasps the paper.
Strategy 2 has the same necessary conditions (Eq. (\ref{eq01})-Eq. (\ref{eq03})) as Strategy 1 and has two grasping scenarios: paper-edge and non-edge graspings (Fig. \ref{graspstrategy}(b) (3) and (4)).
In this strategy, the working parameters include the initial distance $W_0$, opening distance $W_1$, contact width $W$, finger length $L_f$, deformed finger height $H$, and angle of gripper $\theta$. 
Assuming that the finger is in the critical state of contacting the desktop in its natural state, the maximum tilt angle and its corresponding minimum grasping length are as follows:
\begin{equation}\label{add_eq1}
    \theta_{max}=\arctan{\frac{L_f-H}{W_0+(W_1-W_0)/2}}=\arctan{\frac{2(L_f-H)}{W_1+W_0}}
\end{equation}
\begin{equation}\label{add_eq2}
\begin{aligned}
    W_{min}&=\sqrt{(L_f-H)^2+(\frac{W_1-W_0}{2} +W_0)^2}\\
    &=\sqrt{(L_f-H)^2+(\frac{W_1+W_0}{2})^2} 
\end{aligned}
\end{equation}

\subsection{Grasping by Exploiting Extra Constraints}
\label{Grasping Exploiting Extra Constrains}

In addition to using the desktop directly, other constraints (such as walls and table edges) in the environment can assist in grasping. 
As  these constraints are often at a certain distance from the grasped object, sliding is usually necessary.

\begin{figure}[htbp]
  \centering
  \includegraphics[width=0.6\textwidth]{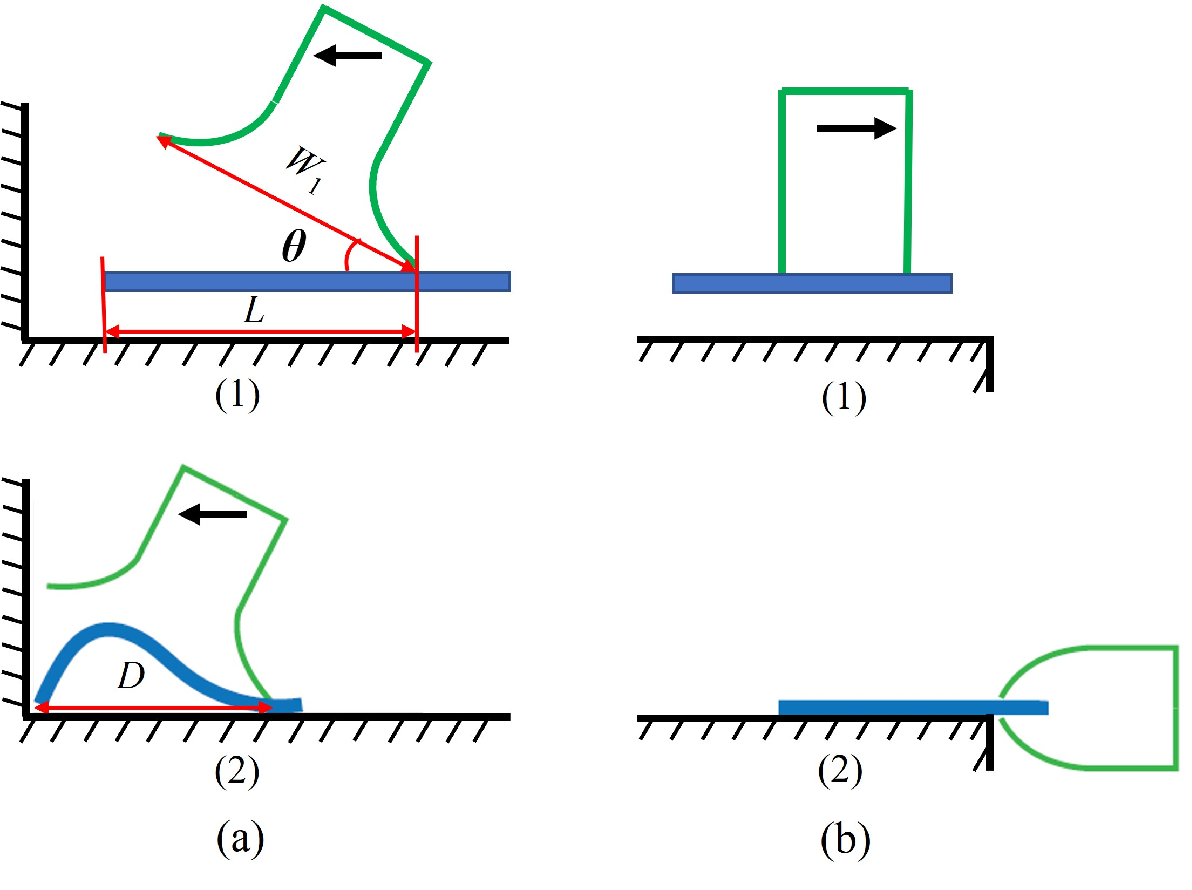}
  \caption{Designing paper-grasping strategies by exploiting extra constraints. (a) Strategy 3: Wall grasp. (b) Strategy 4: Edge grasp. The second row shows the corresponding grasping process of the first row}
  \label{graspstrategy34}
\end{figure}

\textbf{Strategy 3} uses the wall for grasping, as shown in Fig. \ref{graspstrategy34}(a). 
Herein, a wall refers to any object in the environment that can provide contact constraints. 
During grasping, the gripper is open and in contact with the paper on one side. 
It then presses the paper and slides toward the wall. 
When the paper encounters an external wall, it is deformed and then grasped.
As shown in Fig. \ref{graspstrategy34}(a), the working parameters during the grasping process are opening distance $W_1$, angle of gripper $\theta$, wrinkle area $L$, and grasping distance $D$.
In addition to the necessary grasping conditions similar to those of Strategy 1, Strategy 3 applies two conditions for grasping: 1) the gripper does not collide with the wall and 2) the deformation of the paper is within the grasping space of the gripper.
To prevent collisions, the grasping distance $D$ must satisfy Eq. \ref{eq11}.
Because the inclination angle of the gripper changes its grasping space, corresponding angles $\theta$ must be set for different grasping distances $D$ to ensure grasping.
The working space of this strategy was experimentally obtained in Section \ref{Experiment 3: Wall Grasp}.
\begin{equation}\label{eq11}
    W_1 \cdot cos\theta < D < L
\end{equation}

\textbf{Strategy 4} exploits the table edge to grasp the paper (Fig. \ref{graspstrategy34}(b)).
The slide skill is used to push the paper over the edge of the table, and the part beyond the edge is then grasped.
During the sliding stage, the tangential force on the paper must satisfy Eq. (\ref{eq8}); therefore, Eq. (\ref{eq01}) is the necessary condition.
For the excess part of the paper, $L$ may be deformed by gravity and the extent of deformation $R$ must be within the grasping space of the gripper (see Experiment 4 in Section \ref{Experiment 4: edge grasp}).
As the paper does not require active deforming forces, this strategy is also applicable to rigid planar objects.

In this study, the term “workspace” refers to the range of conditions under which a grasping strategy operates successfully. 
To avoid ambiguity, we distinguish between the theoretical workspace, which is derived from geometric or mechanical models (e.g., Eq. (\ref{eq11})), and the experimental workspace, which is determined empirically based on a success rate threshold of 0.6 (see Section \ref{EXPERIMENTAL RESULTS}). 
Strategy-specific workspace characteristics are summarized and compared in Section \ref{DISCUSSION} (Table \ref{table2}).

\section{Experimental Setup}
\label{SETUP FOR EXPERIMENTS}

\subsection{Architecture of Robotic System}

\begin{figure}[htbp]
  \centering
  \includegraphics[width=0.96\textwidth]{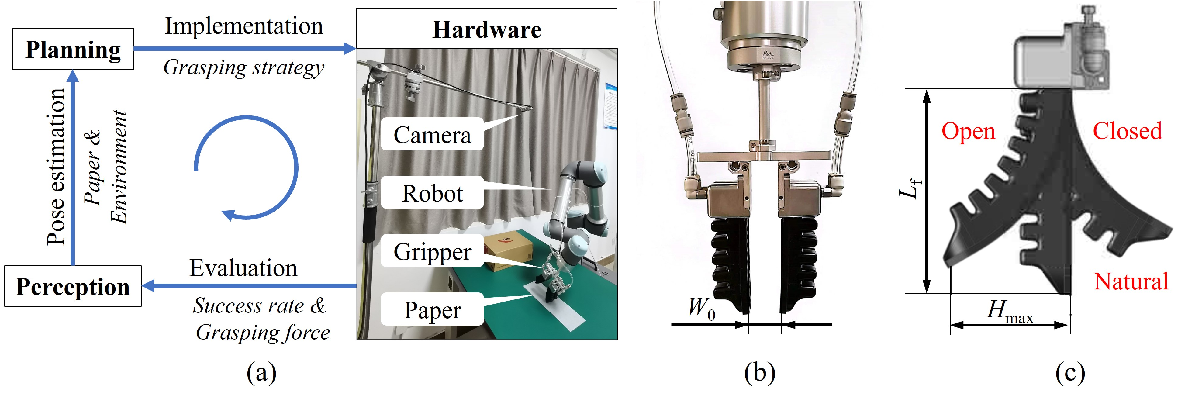}
  \caption{Architecture of robotic system. (c) Software and hardware of the robotic system. (b) Gripper mounted on the end of the robot. (c) Size and workspace of one finger}
  \label{hardware}
\end{figure}

For the experimental equipment, a soft gripper (Rochu GC-4FMA6V5/LS1) was mounted at the end of a robot arm (UR5e, Universal Robots), and a depth camera (RealSense D415, Intel) was fixed 1 m above the table.
The working radius of the robot was 850 mm, and the field of view of the camera at this height was 1.25 m $\times$ 0.74 m.
The dimensions and grasping space of the gripper are shown in Fig. \ref{hardware}.
The initial distance $W_0$ between fingers was 30 mm, length $L_f$ of the finger was 83 mm, and maximum travel $H_{max}$ of the fingertip was 33 mm (-80 kPa).
The gripper operated in three states (open, closed, and natural states), and its opening distance $W_1$ depended on negative pressure.

The ROS  framework was used to integrate hardware, perception, and planning modules (Fig. \ref{hardware} (a)). 
In the perception module, YOLO \cite{wang2021scaled} was used to detect objects, such as the paper to be grasped and other common objects, on the table.
Other objects in the environment provided potential contact constraints to assist in grasping.
Furthermore, following our previous study \cite{dong2023robotic}, their  poses can be calculated through background segmentation and contour fitting.
In addition to object detection, the perception module was also used to obtain the grasping force throughout the entire grasping process. 
The planning module provided different grasping strategies to implement the grasping process.

\subsection{ Strategy Implementation}

\begin{figure}[htbp]
  \centering
  \includegraphics[width=0.96\textwidth]{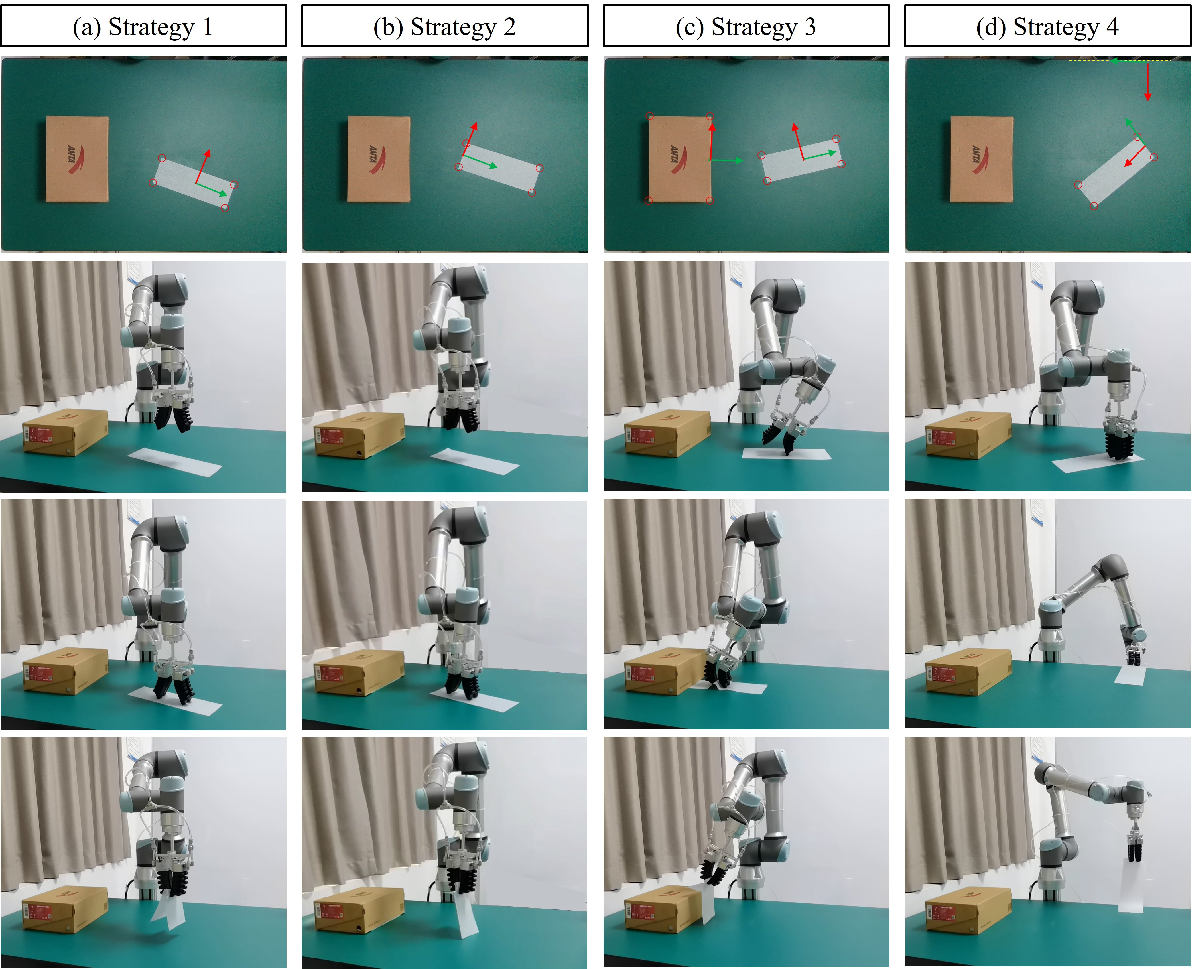}
  \caption{Real robot demonstration of proposed strategies. A video showing these grasping strategies can be seen at \href{https://youtu.be/ta_4K0r8lT4}{\url{https://youtu.be/ta_4K0r8lT4}}}
  \label{Implementation}
\end{figure}

For Strategy 1, the vision module can determine the pose of the paper $P_{paper}$ (Fig. \ref{Implementation} (a)) based on the grasping position $x$ and its corner points $K_{paper}$, which can be obtained via background segmentation and contour fitting \cite{dong2023robotic}.
Based on Strategy 1, the grasp pose of Strategy 2 must consider the tilt angle of the gripper. 
For Strategies 1 and 2, which are based on pinching skills, obtaining the paper pose is sufficient for grasping.

\begin{algorithm}
\caption{Strategy 3 Wall Grasp}\label{algorithm1}
\begin{algorithmic}[1]
\Require{Raw image $Img_{RGBD}$, wrinkle area $L$, gripper angle $\theta$, grasp distance $D$}
\Ensure{Transformation matrix $T$}
\State $(K_{paper}, K_{box}) \gets Img_{RGBD}$
\State $(P_{paper}, P_{box}) \gets (K_{paper}, K_{box})$
\State $S_{gripper} = \text{open}$
\State $P_{initial} \gets \text{initialPose}(P_{paper}, L, \theta)$
\State $P_{objective} \gets \text{objectivePose}(P_{box}, D)$
\State $T \gets \text{transformation}(P_{initial}, P_{objective})$
\State $S_{gripper} = \text{close}$
\State \Return $T$
\end{algorithmic}
\end{algorithm}

The grasp planning goal of Strategy 3 is to calculate the transformation matrix $T$ from the initial slide pose $P_{initial}$ (Fig. \ref{graspstrategy34} (a)(1)) to the grasp pose $P_{objective}$ (Fig. \ref{graspstrategy34} (a)(2)).
The work scene is as shown in Fig. \ref{Implementation}, and the box on the table can provide wall-like contact constraints.
The grasping planning algorithm for Strategy 3 is given by Algorithm \ref{algorithm1}.
First, the keypoints $K_{paper}$ and $K_{box}$ of the paper and box are extracted from RGB data through object recognition, segmentation, and contour fitting processes \cite{dong2023robotic}. 
Then, by combining these visual information with depth data, the poses of the paper $P_{paper}$ and the box $P_{box}$ can be computed, as illustrated in Fig. \ref{Implementation} (c).
Before grasping, based on the paper pose $P_{paper}$, the orientation of the initial pose $P_{initial}$ can be determined using the tilt angle $\theta$, and its position can be calculated based on the wrinkled area $L$.
In addition, the target pose $P_{objective}$ of the gripper is determined by the pose of the box $P_{box}$ and the grasping distance $D$.
During grasping, the gripper slides from the initial pose $P_{initial}$ to $P_{objective}$ and then grasps the wrinkles on the paper.
Similarly, Strategy 4 determines the initial and target poses based on the paper pose and edge of the desktop \cite{bimbo2019exploiting}.
Implementation scenarios for these strategies are shown in Fig. \ref{Implementation}.

\subsection{Experimental Evaluation}
\label{Metrics}

\begin{figure}[htbp]
  \centering
  \includegraphics[width=0.7\textwidth]{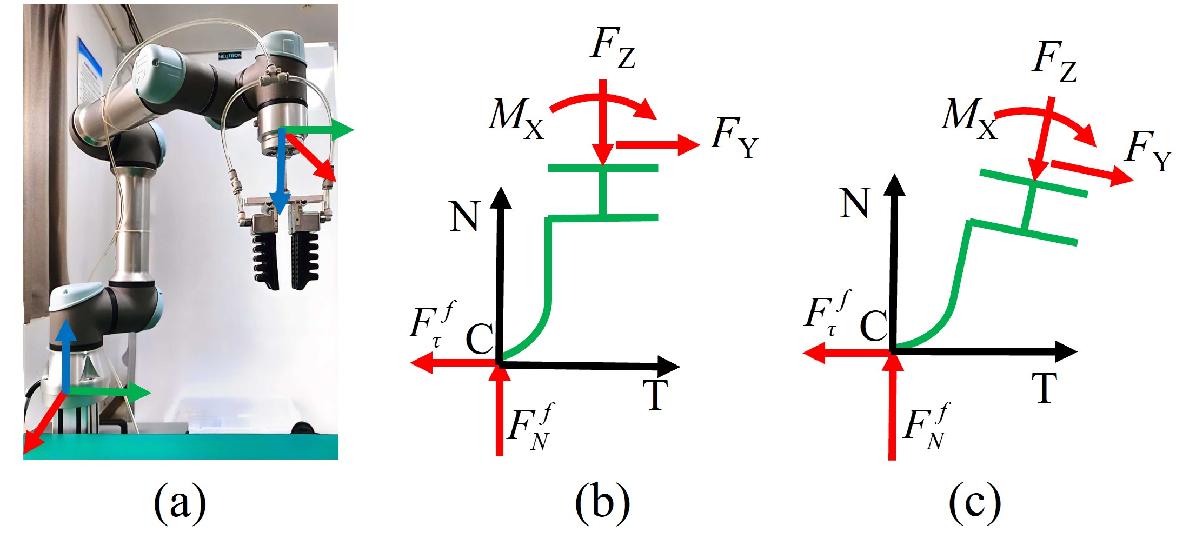}
  \caption{Coordinate system and grasping force analysis. (a) The coordinate system of the robot. (b) Analysis of the grasping force of single-sided fingers in Strategy 1. (c) Analysis of the grasping force of single-sided fingers in Strategies 2 and 3}
  \label{metricsForce}
\end{figure}

To explore the workspace and working characteristics of the different strategies, we tested their grasping forces and success rates under different working conditions.
For each condition, the success rate was determined from ten trials, with each attempt considered successful only if the gripper could securely lift the paper from the table.
To understand the effect of the grasping force on the grasping process, we simultaneously tested the changes in the grasping pose and  force. 
The grasping forces included the grasping force of the entire gripper and that of the single-sided fingers. 
These forces, which had different functions, were measured using a force sensor mounted on the wrist of the robotic arm. 
The grasping force of the entire gripper was used to monitor the grasping process, whereas the grasping force of the fingers was tested separately to analyze the cause of grasping principle. 
The coordinate systems of the robot arm base and end force sensor are shown in Fig. \ref{metricsForce} (a).
The force analysis diagrams of the unilateral fingers for Strategies 1--3  are shown in Fig. \ref{metricsForce} (b-c).
$F_Z$, $F_Y$, and $M_X$, representing the reaction forces measured using the force sensor, are the external forces acting on the end effector at the connection to the robot.
$F_N^f$ and $F_\tau^f$ are the forces acting on the finger contact point. These forces can be obtained using the following equilibrium equations (Eqs. (\ref{fig6_eq1}) \& (\ref{fig6_eq2})). 
In addition, they interact with the normal force $F_P$ and tangential force $F_\tau$ as action and reaction forces, respectively, as shown in Fig. \ref{skillforce} (b).

\begin{equation}\label{fig6_eq1}
\left\{\begin{matrix} 
  -F_Z+F_N^f=0 \\  
  F_Y-F_\tau^f =0 
\end{matrix}\right. 
\end{equation}
\begin{equation}\label{fig6_eq2}
\left\{\begin{matrix} 
  -F_Z\text{cos}\theta-F_Y\text{sin}\theta+F_N^f=0 \\  
  -F_Z\text{sin}\theta+F_Y\text{cos}\theta-F_\tau^f=0 
\end{matrix}\right. 
\end{equation}

\subsection{Specimen}
According to Chinese national standards, different types of paper have specific specification ranges, expressed in grams per square meter (GSM).
Three distinct types of papers were prepared to test the workspaces of the different strategies.
\begin{itemize}
\item Copy paper, GSM = 17 / 35 g.
\item Printing paper, GSM = 60 / 80 / 100 / 120 g.
\item White cardboard, GSM = 150 / 200 / 230 / 250 g.
\end{itemize}
The size of the test paper was uniformly 105 mm × 297 mm, which is half the size of an A4 paper.

\section{Experimental Results and Analysis}
\label{EXPERIMENTAL RESULTS}

\subsection{Experiment 1: Top Grasping}

The experiment for Strategy 1 primarily focused on the grasping effect of the grasping position and paper specifications. 
Following the definition of Strategy 1 (Section \ref{Grasping Exploiting Tabletop}), the grasping position is determined by two parameters: the initial distance $W_0$ and opening distance $W_1$ of the grippers.
The $W_0$ and $W_1$ values of the Rochu soft gripper were 30 and 96 mm, respectively (Fig. \ref{hardware}). Therefore, the grasping position $x$ was set to 50, 70, 90, 110, and 130 mm.

\begin{figure}[htbp]
  \centering
  \includegraphics[width=0.85\textwidth]{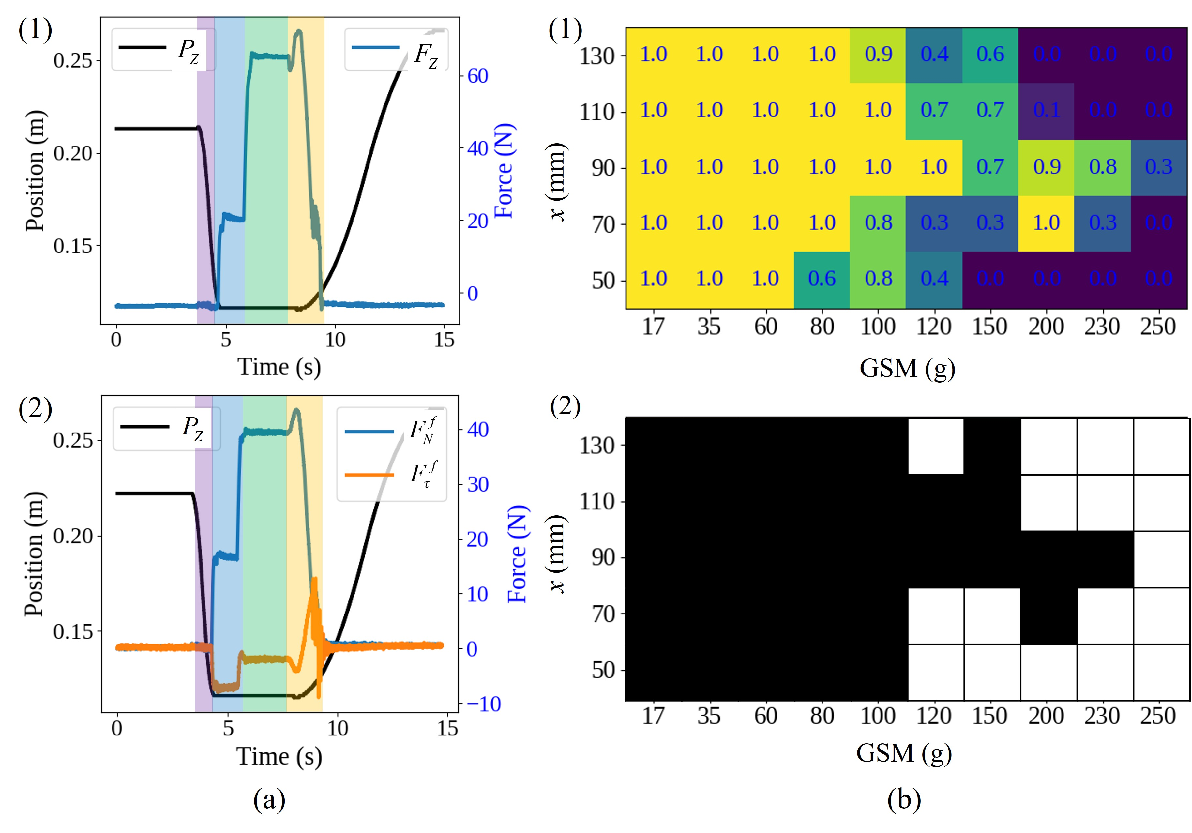}
  \caption{Experimental results of Strategy 1. (a) Grasping position and force of entire gripper and one-sided fingers during grasping under Strategy 1. (b) Success rate and experimental workspace of Strategy 1 under different materials and grasping positions}
  \label{experiments_s1}
\end{figure}

To explore the grasping process in detail, the grasping forces of the gripper and unilateral fingers were tested. The results are shown in Fig. \ref{experiments_s1}(a). 
The force diagram of the gripper (Fig. \ref{experiments_s1}(a) (1)) shows that the grasping process is divided into four stages, namely, the approach, contact, pressure, and lifting stages. 
During grasping, the gripper approaches the paper in an open state (approach stage) until it touches the paper (contact stage). At this time, a positive normal force $F_Z$ acts on the gripper.
Then, the gripper receives a closing command and enters the pressure stage, during which the normal force increases further.
When the gripper moves up (lifting stage), it automatically grasps the paper and the positive pressure is removed. 
As expressed in Eqs. (\ref{eq02}) and (\ref{eq03}), the key factors that determine whether the paper can be grasped are the values of the normal and tangential forces.
Fig. \ref{experiments_s1}(a) (2) shows the normal and tangential forces exerted by the left finger. 
Based on the analysis of the grasping process, among the four stages of grasping, the paper was grasped in the lifting stage. Therefore, we focused on the grasping force value in this stage.
The direction of the tangential force changed to the opposite direction, which is attributable to the tendency of the fingers to close as the gripper traverses upward. 
Its normal force continued to decrease, whereas the tangential force increased to a peak value of over 10 N and then decreased, corresponding to the first and second stages of the pinching skill, balance stage, and movement stage.

The effects of grasping positions and materials on the grasping success rate are shown in Fig. \ref{experiments_s1} (b)(1). 
Regarding paper specifications, the overall trend was that the higher the GSM value, the lower the success rate.
A higher GSM value indicates a thicker paper. The bending stiffness of the paper also increases, and a greater external force is required for the paper to yield (Eq. (\ref{eq02})).
When the grasping position is close to the edge of the paper ($x =90$ mm), the grasping success rate is the highest overall. 
In particular, when the paper specification GSM $ > $ 100 g, this edge effect is significant.
The edge effect is caused by two main factors: the small paper-deformation force at this time (with a long grasping length $L$ and large length coefficient $\mu$) (Eq. (\ref{eq02})) and the edge shape factor.
To obtain the working space of Strategy 1, the threshold of the success rate is set to 0.6, and its experimental workspace distribution diagram is shown in Fig. \ref{experiments_s1}(b) (2).
Near the edge of the paper ($x = 90$ mm), the paper specification that can be grasped is 230 g.

\subsection{Experiment 2: Top Scooping}

Compared with Strategy 1, Strategy 2 introduces a tilt angle during grasping to achieve scooping of paper.
The tilt angle range of the Rochu gripper is approximately $0-7.5^\circ$. During the test, the angle was set to $5^\circ$, and the corresponding contact width $W$ was approximately 80 mm.
In order to validate the edge effect during grasping, an 80-mm condition was added based on the test position of Strategy 1.

\begin{figure*}[htbp]
  \centering
  \includegraphics[width=13cm]{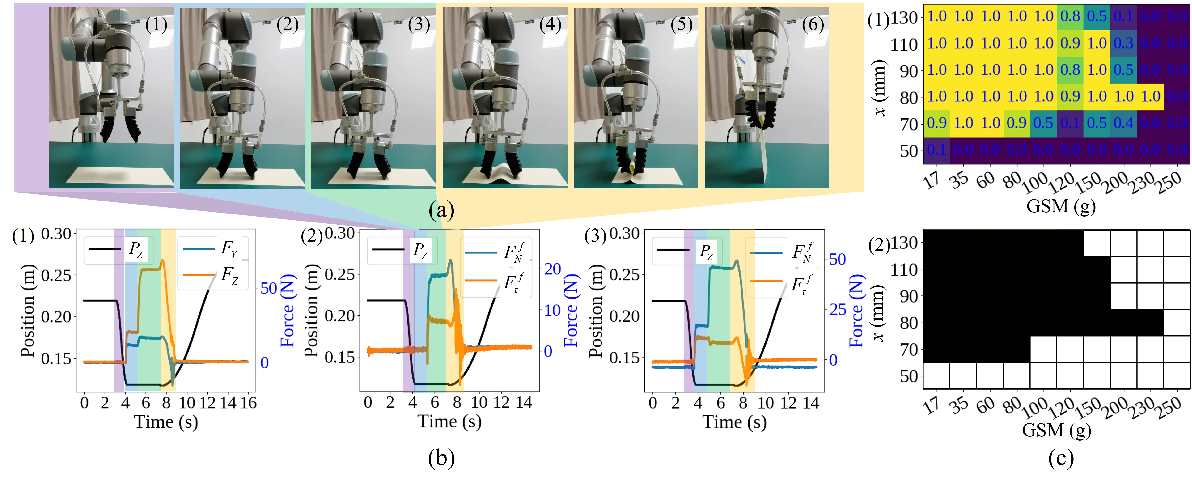}
  \caption{Experimental results of Strategy 2. (a) Complete grasping process of Strategy 2 on actual robot. (b) Grasping position and force during grasping under Strategy 2: (1) Entire gripper, (2) left fingers, and (3) right fingers. (c) Success rate and experimental workspace of Strategy 2 under different materials and grasping lengths}
  \label{experiments_s2}
\end{figure*}

As shown in Fig. \ref{experiments_s2}, the grasping process of Strategy 2 also comprises four stages.
Fig. \ref{experiments_s2}(b) (1) shows the force process of the gripper during grasping.
Compared with that of Strategy 1, the gripper was also subjected to lateral forces during the grasping process as a result of the inclination angle (Fig. \ref{experiments_s2} (b)(1)).
To investigate the effect of the unilateral fingers on grasping, we analyzed the changes in the grasping forces on the left and right fingers (Fig. \ref{experiments_s2}(b) (2) and (3)). 
The left and right fingers exhibited different movement patterns.
The left finger contacted the paper during the pressure stage, and the generated tangential force acted in the same direction until the lifting stage.
The movement pattern of the right finger was similar to that of the finger in Strategy 1, and the tangential force generated changed direction during the lifting phase.

The distribution of the grasping success rates for Strategy 2 is shown in Fig. \ref{experiments_s2}(c).
Similar to that in Strategy 1, the general rule is that the larger the GSM, the greater the probability of failure. 
A larger GSM value indicates a thicker paper, rendering the paper stiffer and consequently reducing the success rate. 
However, its working space differs from that of Strategy 1, particularly when the grasping position $x <$ 80 mm. For example, when $x$ = 50 mm, Strategy 2 fails to grasp the paper because the contact width $W$ of the gripper in this state is greater than 50 mm.
When the grasping position $x >$ 80 mm, Strategy 2 can be applied to higher specifications of the paper than Strategy 1, owing to the special movement pattern of the left fingers.
The edge effect also occurs in Strategy 2. The experimental workspace of Strategy 2 divided in terms of the success rate threshold of 0.6 is shown in Fig. \ref{experiments_s2}(c) (2).

\subsection{Experiment 3: Wall Grasp}
\label{Experiment 3: Wall Grasp}

\begin{figure*}[htbp]
  \centering
  \includegraphics[width=13cm]{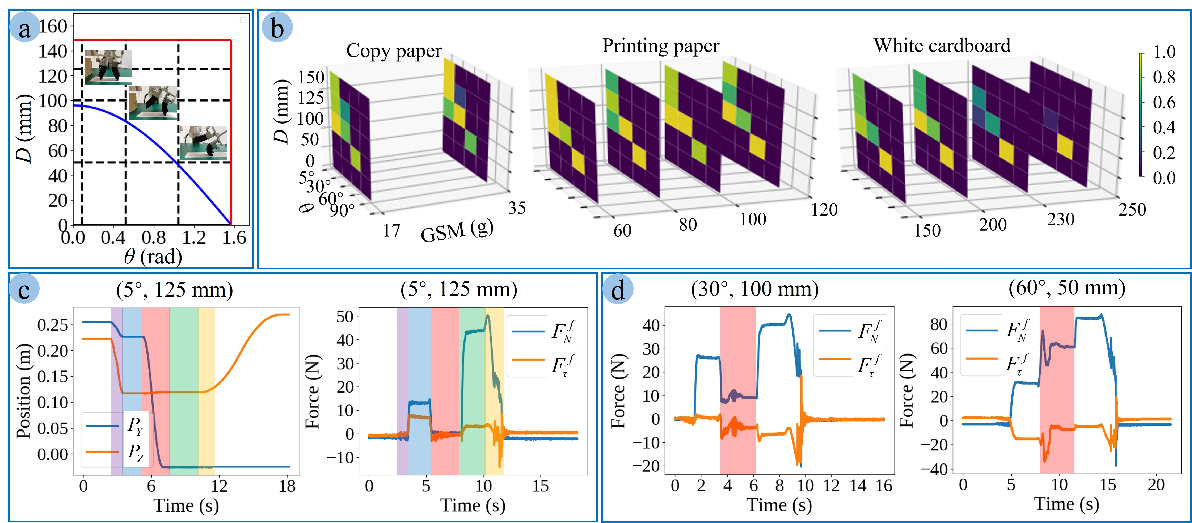}
  \caption{Experimental results of Strategy 3. (a) Theoretical workspace of Strategy 3. (b) Success-rate distribution and experimental workspace of Strategy 3. (c) Grasping position and force during grasping at ($5^\circ$, 125). (d) Grasping force during grasping at ($30^\circ$, 100) and ($60^\circ$, 50)}
  \label{experiments_s3}
\end{figure*}

Strategy 3 focuses on the effects of material, gripper inclination angle $\theta$, and grasping distance $D$ on grasping.
Based on the strategy definition in Section \ref{Grasping Exploiting Extra Constrains} (Eq. \ref{eq11}), the theoretical workspace of Strategy 3 is shown in Fig. \ref{experiments_s3}(a).
During the experiment, the value of $W_1$ was 96 mm, and the initial contact point was set at the center of the paper ($L = 148.5$ mm).
The grasping distance $D$ was set to 0, 50, 100, 125, and 150 mm, and the tilt angle $\theta$ was set to $5^\circ$, $30^\circ$, $60^\circ$, and $90^\circ$.
Additionally, when the inclination angle of the gripper was $90^\circ$, part of the gripper collided with the tabletop and the success rate was set to zero.

To investigate the work characteristics of Strategy 3, we analyzed the influence of working conditions on the grasping force. 
As shown in Fig. \ref{experiments_s3}(c), compared with Strategies 1 and 2, Strategy 3 has an additional slide stage (red color). 
Following the definition of Strategy 3, considerable deformation of the paper during the sliding stage is a necessary condition for grasping.
Therefore, the grasping force during the sliding stage should be investigated.
Fig. \ref{experiments_s3} (c) and (d) show the normal and tangential forces acting in the three working conditions of ($5^\circ$, 125 mm), ($30^\circ$, 100 mm), and ($60^\circ$, 50 mm), where the normal force corresponding to the slide stage increases in sequence.
These normal forces, which were stable at approximately 1, 10, and 60 N, exhibited significant differences.
The main reason for this phenomenon is that the deformation of the fingers differs when sliding at different inclination angles, as shown in the subfigure of Fig. \ref{experiments_s3}(a).

The distribution of success rates for papers with different specifications is shown in Fig. \ref{experiments_s3}(b), which also shows the experimental workspace.
The experimental results indicate that, when the inclination angle $\theta$ of the gripper increases, the grasping distance $D$ must be reduced to achieve grasping. 
For example, when the inclination angle is $60^\circ$, the grasping distance must be reduced to 50 mm.
As discussed in Section \ref{Grasping Exploiting Extra Constrains}, after the bulge was formed in the slide phase, the key condition for the final grasping was that the bulge should be in the grasping space of the gripper.
When the GSM was $\leq$ 200 g, the experimental workspace were similar. 
Except for a few failures due to visual accuracy, the success rate was stable within the range of $70\%-100\%$.
However, when the GSM $\geq$ 230 g, the working space gradually shrank because an increase in paper thickness led to an increase in stiffness; thus, the paper did not deform or form bulges during the sliding stage.
A special case is ($60^\circ$, 50 mm), where the experimental success rate was maintained at a level close to $100\%$.
As the normal force exerted by the gripper in the sliding stage was approximately 60 N when the tilt angle was $60^\circ$, this force  could increase the upper limit of the tangential force used to resist the deformation force of the paper (Eq. (\ref{eq02})) and prevent slippage between the finger and paper (Eq. (\ref{eq03})).

\subsection{Experiment 4: Edge Grasp}
\label{Experiment 4: edge grasp}

\begin{figure}[htbp]
  \centering
  \includegraphics[width=8cm]{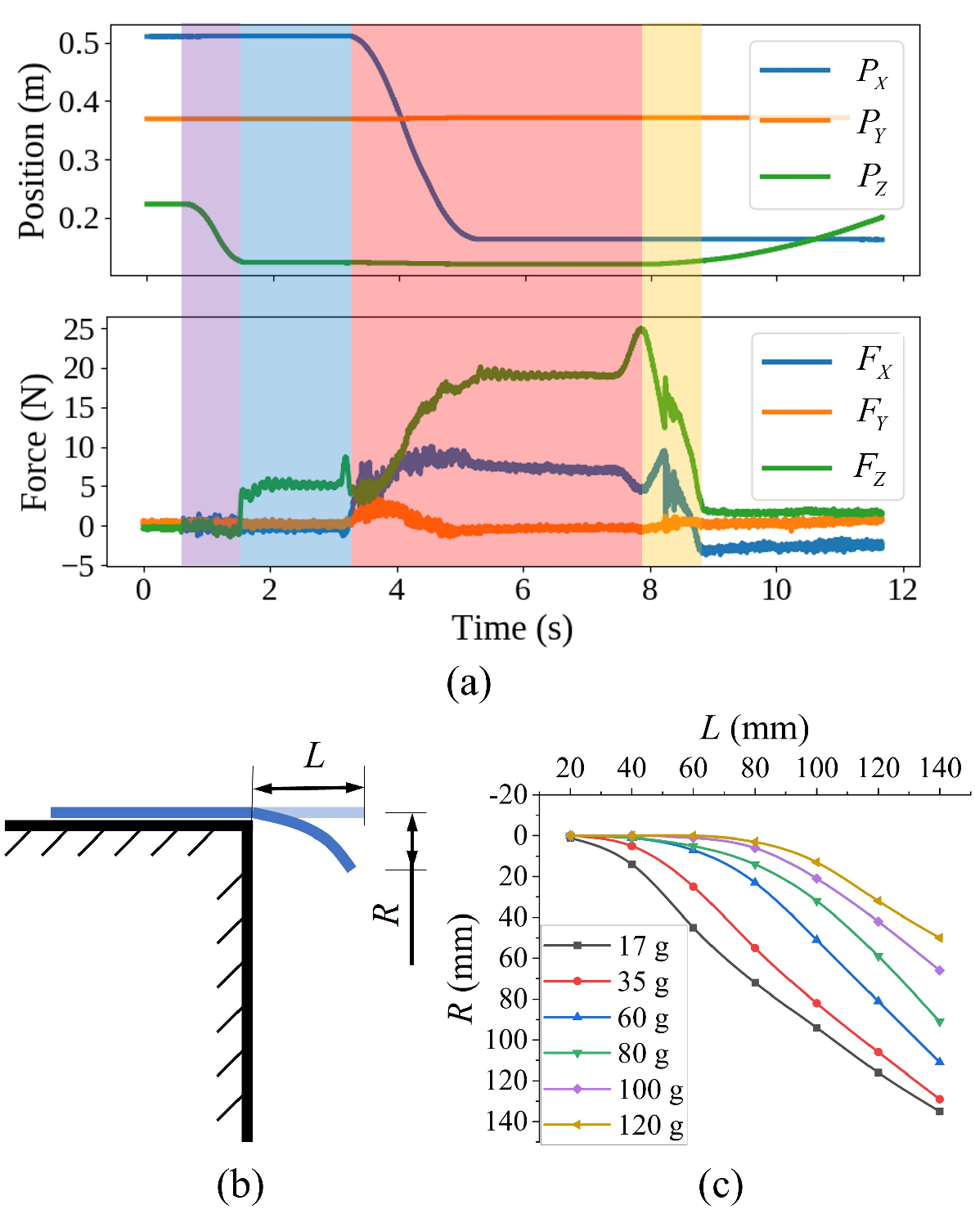}
  \caption{Experimental results of Strategy 4. (a) Position and contact force of the gripper during the sliding process in Strategy 4. (b) Deformation diagram of Strategy 4. (c) Relationship between the end deformation and protruding length under different paper specifications}
  \label{experiments_s4}
\end{figure}

The changes in the position and contact force of the gripper during the sliding process in Strategy 4 are shown in Fig. \ref{experiments_s4}(a).
Strategy 4 comprises four stages: approach, contact, sliding, and lifting. 
Unlike Strategy 3, which satisfies forced deformation in the sliding stage, Strategy 4 needs to satisfy only the condition of paper slippage: $\mu_0 > \mu_1$.
The contact-force fluctuations shown in Fig. \ref{experiments_s4}(a) appear during the sliding phase. 
Because the final grasp is performed in the nonslip stage and targets only the overhanging paper section, Strategy 4 requires sliding to push the paper beyond the table edge. 
Force fluctuations during sliding are irrelevant to the final grasping performance. 
To ensure a stable and nondestructive manipulation, the intrinsic compliance of the soft gripper limits the contact force, and position-based control with a low speed is adopted to ensure a safe and gentle operation. 
Although closed-loop force control is not required, it can be integrated in future studies to enhance robustness.

In Strategy 4, the protruding part must be determined for grasping.
For paper-like flexible materials, the gravity of the protruding part $L$ causes the end of the paper material to sag, assuming that the vertical displacement is $R$ (Fig. \ref{experiments_s4}(b)).
Based on the grasping model in Strategy 4 (Figs. \ref{graspstrategy34}(b)), the tool center point is aligned with the tabletop. 
Therefore, the determination of the workspace is a deterministic geometric problem. It is sufficient to ensure that the deformation of the paper remains within the grasping space of the gripper, that is, $R < W_1/2$, without the need for success rate statistical experiments.
The relationship between end deformation and protrusion length has been tested in studies with different specifications (17 g $\leq$ GSM $\leq$ 120 g), as shown in Fig. \ref{experiments_s4}(c).
When using the Rochu gripper in the experiment for these tested materials, all grasping can be achieved when the protrusion $L \leq$ 60 mm.

The relationship shown in Fig. \ref{experiments_s4}(c) is specific to the tested GSM range (17--120 g) and is not directly generalizable to untested materials, as the deformation depends on the bending stiffness $EI \propto E \cdot h^3$.
To extend the results, we propose performing normalization based on $EI$.
The cantilever deflection is expressed as $R = \frac{\lambda g L^4}{8EI}$, where $\lambda$ denotes the linear mass density of the paper strip. 
As a key indicator in the experiments, the GSM represents the areal density of the paper, which is positively correlated with $\lambda$ and can be mutually converted based on the paper width.
This formula provides a scaling law: for a specified material, the required protrusion length $L$ is scaled to $(EI)^{1/4}$ to achieve the same end-deformation $R$.
In practice, a simple cantilever test can measure the $EI$ of a new material, thus enabling the robot to estimate the appropriate $L$ without exhaustive retesting.

\section{Discussion}
\label{DISCUSSION}

The analysis of the experimental results shows that the grasping process of these strategies comprises multiple stages, among which the lifting stages of Strategies 1 and 2 and the sliding stage of Strategy 3 determine the success or failure of grasping. 
At this critical stage, if the normal and tangential forces satisfy Eqs. (\ref{eq02}) and (\ref{eq03}), the paper can form wrinkles for grasping.
Strategies 1 and 2 satisfy the requirements of common paper-like materials (GSM $\leq$ 230 g). 
Both strategies have noticeable edge effects because the deformation force generated by the paper is the smallest at this time, and a shape factor is observed.
Owing to the scooping movement mode of the fingers, Strategy 2 offers  advantages different to those of Strategy 1: when $x > W$ (non-edge grasping), Strategy 2 features a larger workspace, and vice versa.
Similar to Strategies 1 and 2, Strategy 3 also involves forced deformation of paper; however, the large deformation of the fingers caused by the large inclination angle of this strategy in the slide stage enables it to provide a large normal force and thus grasp thick papers (GSM $\geq$ 250 g).
As discussed in Section \ref{Grasping Exploiting Extra Constrains}, in addition to satisfying the conditions required for paper deformation (Eqs. (\ref{eq02}) and (\ref{eq03})), it is necessary to consider collision avoidance and whether the paper ridge is in the grasping space of the gripper. Consequently, the working space of Strategy 3 is complex (as shown in Fig. \ref{experiments_s3}(b)).
In contrast to the previous strategies, Strategy 4 does not involve forced deformation of the paper. Therefore, its grasping needs to satisfy only the friction coefficient condition in Eq. (\ref{eq01}). 
Moreover, for Strategy 4, increasing the paper GSM reduces passive deformation, thereby facilitating the grasping process. 
Consequently, this strategy is applicable not only to flexible materials with high GSM but also to planar rigid objects.

In addition to the workspace, these strategies have different working characteristics that are crucial for their application (Table \ref{table2}).
These working characteristics include the size of the object, surface quality, and controllability of placement.
The main requirement for the size of an object is whether it can exceed the workspace of the robot.
Strategies 1 and 2 can grasp the object directly, whereas Strategies 3 and 4 require that the object and environmental constraints be in the workspace of the robot.
The evaluation of surface quality is mainly determined by the length of the crease on the surface of the paper after grasping, and surface quality is important for tasks such as printing, writing, and drawing.
Strategies 1 and 2 involve deformation and squeezing of the paper; thus, creases are formed on the surface of the paper, especially for thin papers \cite{dong2023robotic}.
The controllability of placement mainly refers to the accuracy of the placement pose. For edge grasping, the placement pose of the paper can be controlled. For non-edge grasping, as in Strategy 3, it is uncontrollable.

\begin{table}[ht]
\caption{The workspace and working characteristics of the strategies and their potential application cases}
\label{table2}
\centering
{\fontsize{5.5}{6}\selectfont
\renewcommand{\tabcolsep}{2pt}
\begin{tabular}{lccccccccc}
\toprule
\multirow{3}{*}{Strategy} 
& \multirow{3}{*}{\makecell[c]{Equivalent\\stiffness\\(maximum GSM)}}
& \multirow{3}{*}{Size}   
& \multicolumn{2}{c}{Task requirement} 
& \multicolumn{5}{c}{Potential application} \\
\cmidrule(lr){4-5}
\cmidrule(lr){6-10}
& & & \makecell[c]{Surface\\quality}
& \makecell[c]{Place pose\\controllability}
& \makecell[c]{Printing\\paper}
& \makecell[c]{Packaging\\paper}
& Cardboard 
& Tablecloth 
& Tissue \\
\midrule
Strategy 1 
& 230g
& All size 
& \makecell[c]{Not good}
& Edge grasp
& $\times$ & \checkmark & $\times$ & \checkmark & \checkmark \\
Strategy 2 
& 230g 
& All size
& \makecell[c]{Not good}
& Edge grasp
& $\times$ & $\times$ & $\times$ & \checkmark & $\times$ \\
Strategy 3 
& $\geq$250g 
& \makecell[c]{In the robot's\\workspace}
& Good 
& No 
& $\times$ & $\times$ & $\times$ & $\times$ & $\times$ \\
Strategy 4 
& All
& \makecell[c]{In the robot's\\workspace} 
& Best 
& Yes 
& \checkmark & $\times$ & \checkmark & $\times$ & $\times$ \\
\bottomrule
\end{tabular}}
\end{table}

In specific applications, an appropriate strategy can be selected based on the physical properties of the object and task requirements.
Some potential applications are presented in Table \ref{table2}.
The use of printing paper (GSM = 70--80 g) not only requires high surface quality but also demands controllability of the placement pose, such as inserting the paper into a printer.
Therefore, Strategy 4 is the most suitable among the proposed strategies for this application.
When grasping packaging paper (GSM = 17 g), the edge-grasping method can control the pose of the paper to facilitate downstream operational tasks \cite{dong2023robotic}. At this time, Strategy 1 is preferred, as it is more convenient and faster, without the need to adjust the gripper’s tilt angle as required in Strategy 2.
For cardboard, which typically has a much larger GSM value, grasping is similar to grasping rigid planar objects, and Strategy 4 can be used directly for grasping \cite{bimbo2019exploiting}.

The proposed strategies are mechanically generalizable to other two-dimensional deformable materials, such as tablecloths and tissues, although no additional experiments were conducted on these materials in this study. 
The abovementioned materials share similar features with paper (low bending stiffness and planar deformability); therefore, the same grasping principles apply. 
For tablecloths with extremely low stiffness and large sizes that may exceed the robot workspace, both Strategies 1 and 2 are applicable. Notably, Strategy 2 offers a larger workspace for non-edge grasping and can be used as an alternative to Strategy 1.
For extremely soft tissues with no surface-quality requirements, pinch-based Strategies 1 and 2 function directly, with Strategy 1 preferred for its simpler operation without an additional tilt angle. 
Because these materials have a lower stiffness and more favorable friction than paper, they are easier to grasp reliably. 
Future studies shall include the experimental validation of fabrics and films.

\begin{table}[htbp]
\caption{Comparison with existing grasping methods for flat objects}
\label{Comparison table}
\centering
\footnotesize
\renewcommand{\tabcolsep}{2pt}
\begin{tabular}{lccccccc}
\toprule
\makecell[c]{Paper\\manipulation} 
& Task 
& \makecell[c]{End\\effector} 
& \multicolumn{3}{c}{Environment constrains} 
& \multicolumn{2}{c}{Flat objects} \\
\cmidrule(lr){4-6} \cmidrule(lr){7-8}
& & & \makecell[c]{Table\\contact} & Wall & Edge & Rigid & Flexible \\
\midrule
This study 
& Grasping 
& Universal soft gripper
& \includegraphics[width=0.8em]{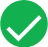}
& \includegraphics[width=0.8em]{label_true.eps}
& \includegraphics[width=0.8em]{label_true.eps}
& \includegraphics[width=0.8em]{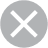}
& \includegraphics[width=0.8em]{label_true.eps} \\
Jiang et al. \cite{jiang2024rotipbot}
& Singulating
& Rotatable fingertips
& \includegraphics[width=0.8em]{label_true.eps}
& \includegraphics[width=0.8em]{label_false.eps}
& \includegraphics[width=0.8em]{label_false.eps}
& \includegraphics[width=0.8em]{label_false.eps}
& \includegraphics[width=0.8em]{label_true.eps} \\
Jiang et al. \cite{jiang2019dynamic, zhao2023flipbot}
& Flipping
& Printed soft gripper
& \includegraphics[width=0.8em]{label_true.eps}
& \includegraphics[width=0.8em]{label_false.eps}
& \includegraphics[width=0.8em]{label_false.eps}
& \includegraphics[width=0.8em]{label_false.eps}
& \includegraphics[width=0.8em]{label_true.eps} \\
Xiong et al. \cite{xiong2023rapid}
& Grasping
& Bistable gripper
& \includegraphics[width=0.8em]{label_true.eps}
& \includegraphics[width=0.8em]{label_false.eps}
& \includegraphics[width=0.8em]{label_false.eps}
& \includegraphics[width=0.8em]{label_false.eps}
& \includegraphics[width=0.8em]{label_true.eps} \\
Zhang et al. \cite{zhang2022prying}
& Prying
& \makecell[c]{Rigid two-fingered\\gripper}
& \includegraphics[width=0.8em]{label_true.eps}
& \includegraphics[width=0.8em]{label_false.eps}
& \includegraphics[width=0.8em]{label_false.eps}
& \includegraphics[width=0.8em]{label_true.eps}
& \includegraphics[width=0.8em]{label_false.eps} \\
Turco et al. \cite{turco2021grasp}
& Grasping
& \makecell[c]{Soft reconfigurable\\gripper}
& \includegraphics[width=0.8em]{label_true.eps}
& \includegraphics[width=0.8em]{label_true.eps}
& \includegraphics[width=0.8em]{label_false.eps}
& \includegraphics[width=0.8em]{label_true.eps}
& \includegraphics[width=0.8em]{label_false.eps} \\
Yuan et al. \cite{yuan2020design}
& Grasping
& Roller-Based hand
& \includegraphics[width=0.8em]{label_true.eps}
& \includegraphics[width=0.8em]{label_false.eps}
& \includegraphics[width=0.8em]{label_false.eps}
& \includegraphics[width=0.8em]{label_true.eps}
& \includegraphics[width=0.8em]{label_true.eps} \\
Morino et al. \cite{morino2020sheet}
& Grasping
& Sheet-based gripper
& \includegraphics[width=0.8em]{label_true.eps}
& \includegraphics[width=0.8em]{label_false.eps}
& \includegraphics[width=0.8em]{label_false.eps}
& \includegraphics[width=0.8em]{label_true.eps}
& \includegraphics[width=0.8em]{label_true.eps} \\
Ko \cite{ko2020tendon}
& Grasping
& \makecell[c]{Tendon-Driven\\Gripper}
& \includegraphics[width=0.8em]{label_true.eps}
& \includegraphics[width=0.8em]{label_false.eps}
& \includegraphics[width=0.8em]{label_false.eps}
& \includegraphics[width=0.8em]{label_true.eps}
& \includegraphics[width=0.8em]{label_true.eps} \\
Babin et al. \cite{babin2019stable, babin2018picking}
& Scooping
& \makecell[c]{Passive and epicyclic\\mechanisms}
& \includegraphics[width=0.8em]{label_true.eps}
& \includegraphics[width=0.8em]{label_false.eps}
& \includegraphics[width=0.8em]{label_false.eps}
& \includegraphics[width=0.8em]{label_true.eps}
& \includegraphics[width=0.8em]{label_true.eps} \\
Hang et al. \cite{hang2019pre}
& Grasping
& \makecell[c]{Soft and compliant\\hands}
& \includegraphics[width=0.8em]{label_true.eps}
& \includegraphics[width=0.8em]{label_false.eps}
& \includegraphics[width=0.8em]{label_true.eps}
& \includegraphics[width=0.8em]{label_true.eps}
& \includegraphics[width=0.8em]{label_false.eps} \\
Bimbo et al. \cite{bimbo2019exploiting}
& Grasping
& Compliant hand
& \includegraphics[width=0.8em]{label_true.eps}
& \includegraphics[width=0.8em]{label_false.eps}
& \includegraphics[width=0.8em]{label_true.eps}
& \includegraphics[width=0.8em]{label_true.eps}
& \includegraphics[width=0.8em]{label_true.eps} \\
Sarantopoulos et al. \cite{sarantopoulos2018human}
& Grasping
& \makecell[c]{Rigid three-finger\\gripper}
& \includegraphics[width=0.8em]{label_true.eps}
& \includegraphics[width=0.8em]{label_false.eps}
& \includegraphics[width=0.8em]{label_true.eps}
& \includegraphics[width=0.8em]{label_true.eps}
& \includegraphics[width=0.8em]{label_false.eps} \\
Odhner et al. \cite{odhner2013open, odhner2012precision}
& Grasping
& \makecell[c]{Underactuated\\hands}
& \includegraphics[width=0.8em]{label_true.eps}
& \includegraphics[width=0.8em]{label_false.eps}
& \includegraphics[width=0.8em]{label_false.eps}
& \includegraphics[width=0.8em]{label_true.eps}
& \includegraphics[width=0.8em]{label_false.eps} \\
Kristek et al. \cite{kristek2012orienting}
& Orienting
& \makecell[c]{A pair of robot\\manipulators}
& \includegraphics[width=0.8em]{label_true.eps}
& \includegraphics[width=0.8em]{label_true.eps}
& \includegraphics[width=0.8em]{label_false.eps}
& \includegraphics[width=0.8em]{label_false.eps}
& \includegraphics[width=0.8em]{label_true.eps} \\
Yoshimi et al. \cite{yoshimi2012picking}
& Grasping
& \makecell[c]{Soft gripper with\\fingertips}
& \includegraphics[width=0.8em]{label_true.eps}
& \includegraphics[width=0.8em]{label_false.eps}
& \includegraphics[width=0.8em]{label_false.eps}
& \includegraphics[width=0.8em]{label_true.eps}
& \includegraphics[width=0.8em]{label_true.eps} \\
\bottomrule
\end{tabular}
\end{table}

Table \ref{Comparison table} compares all existing grasping methods for flat thin-like objects (including rigid and flexible objects) in chronological order.
In 2012, Takashi et al. \cite{yoshimi2012picking} imitated the human grasping method and developed a gripper that uses nails for grasping planar rigid and flexible objects. 
Subsequently, environmental constraints have attracted considerable attention \cite{kristek2012orienting, sarantopoulos2018human, hang2019pre}. The use of environmental constraints (such as edges and walls) can simplify the complexity of grippers and compensate for errors in pose estimation.
In addition to environmental constraints, special manipulation primitives \cite{babin2019stable, babin2018picking, zhang2022prying} and grippers with movable surfaces \cite{yuan2020design, morino2020sheet, ko2020tendon} have been developed for grasping flat objects.
These methods were developed mainly for rigid planar objects. 
Although some methods can also be applied to flexible objects, they \cite{yuan2020design, morino2020sheet, ko2020tendon, babin2019stable, babin2018picking} rely on special grippers. Moreover, these works mainly report functional demonstrations of flexible-object grasping, without an in-depth investigation of the underlying grasping principles.
In recent years, grasping methods for thin and flexible objects have been extensively investigated, focusing on special end effectors (such as rotatable fingertips \cite{jiang2024rotipbot}, bistable grippers \cite{xiong2023rapid}), and dynamic grasping strategies \cite{zhao2023flipbot, jiang2019dynamic}.
Based on grasping methods for rigid planar objects, environmental constraints can also assist in the grasping of thin flexible objects using only a single soft gripper. However, this statement has not been systematically investigated.
This study investigated the use of environmental constraints to assist the grasping of paper-like materials with a universal soft gripper.

A comprehensive comparison between the proposed environmental-constraint-based strategies and existing grasping methods for paper-like flexible materials \cite{jiang2024rotipbot, xiong2023rapid, zhao2023flipbot, jiang2019dynamic} is provided as follows: 
In terms of efficiency, the proposed method uses a universal commercial soft gripper without customized structures or complex control, thus resulting in higher efficiency and easier deployment. 
In terms of robustness, leveraging environmental constraints reduces the sensitivity to positioning errors and enhances the adaptability to varying material stiffness, thickness, and surface conditions. 
In terms of success rate, the proposed strategies offer stable grasping performance across a wide range of paper specifications (17–250 g). Strategy 3 supports papers up to $\geq$ 250 g, while Strategy 4 achieves almost $100\%$ success under the appropriate geometric conditions. 
In general, the proposed method offers superior efficiency, robustness, and success rate under diverse material and operating conditions, thus rendering it more suitable for unstructured household and service-robot scenarios. 

Next, we present the limitations of the mechanical and kinematic models proposed in this study. 
These models assume that paper behaves as a homogeneous, linearly elastic beam whose friction coefficients remain constant. 
However, actual paper-like materials may exhibit anisotropic stiffness, varying thicknesses, non-uniform elasticity, or evolving surface frictions. 
These factors can cause discrepancies between model predictions and the actual deformation or sliding performance. 
For household service robots operating in unstructured environments, such limitations may reduce the grasping stability when unknown or varied materials are involved. 
Future studies shall focus on adaptive model updating using real-time sensing to mitigate these effects. 

The soft gripper used in this study is a commercial universal gripper from Rochu Robotics Company, Ltd. and not a specially designed gripper; therefore, the experimental results are completely reproducible and have direct application value.
The characteristics of these strategies are presented in Table \ref{table2}.
The qualitative working characteristics in the table are universal; however, the quantitative working space depends on the specific structure and size of the gripper. 
Although the working space of a customized soft gripper must be tested separately, the experimental method proposed in this study can still be employed.

\section{Conclusion}
\label{CONCLUSION}

Inspired by human manipulation primitives, this study first defines the manipulation skills for soft grippers and then proposes a series of grasping strategies that exploit environmental constraints for paper-like flexible materials. The grasping principles and corresponding models of these strategies are systematically derived.
To investigate the working characteristics of these strategies, an evaluation system that considered success rate and grasping force was defined.
Experiments were conducted using a real robot to investigate the influence of working conditions and materials on the grasping effect. 

Next, we summarize the key quantitative results of the proposed strategies. 
For Strategy 1 (top grasping) and Strategy 2 (top scooping), the grasping success rate exceeds $60\%$ for paper up to 230 g, with the highest success rate reaching $100\%$ at the optimal edge-grasping position. 
The peak tangential grasping force during the lifting stage can exceed 10 N, which is sufficient to induce paper buckling for stable grasping.
 Strategy 3 (wall grasp) achieves a success rate of $70\%-100\%$ for paper under 200 g and can accommodate paper of 250 g or more under suitable inclination angles, with the normal force during sliding maintained at 1--60 N under typical operating conditions. 
 Strategy 4 (edge grasp) realizes reliable grasping for all tested paper specifications (17--120 g) when the overhang length $L \leq 60$ mm, with a success rate of approximately $100\%$ and no forced deformation required.

The final experimental results provided the workspace, characteristics of the different strategies, and potential application examples. 
This study introduced environmental constraints into the grasping of paper-like materials using a universal gripper and systematically investigated the process, principles, workspace, and working characteristics of these grasping strategies.
Our research result provides a reference for household or commercial service robots for grasping flexible materials.

In the future, for specific grasping strategies, we will focus on adaptive grasping based on tactile sensing, which can further improve the grasping success rate via slip feedback and contact force control.
Additionally, when applied, reasonable pre-grasp planning should be provided for specific scenarios and tasks.

\backmatter





\bmhead{Acknowledgements}

We also would like to give heartfelt thanks to Yangjun Liu, Guangyuan Zang and Donghao Shao who gave comments in this paper!

\section*{Declarations}

\begin{itemize}
\item Ethics approval and consent to participate: Not applicable.
\item Consent for publication: Not applicable.
\item Availability of data and material: The data and material are available from the authors upon reasonable request.
\item Funding: This work was supported by the National Natural Science Foundation of China (NSFC) under Grant Nos. 52205017 and 62233008, and by the Aviation Foundation under Grant No. 2022Z050052001.
\item Conflict of Interest: The authors declare that they have no conflict of
interest.
\item Authors' contributions: Yi Dong conceived the study, designed the methodology, conducted the experiments, performed data analysis, and developed the software. He also prepared the original draft of the manuscript and carried out the visualization and validation of the results. Yang Li provided supervision and research resources. Jinjun Duan supervised the project and acquired the funding. Zhendong Dai contributed research resources. All authors reviewed and approved the final manuscript.
\end{itemize}

\end{document}